\documentclass[12pt]{article}

\usepackage{amsmath,color,amsthm,amssymb,cite,pdfpages,setspace}
\usepackage{graphicx}
\usepackage{hyperref}
\usepackage{caption}
\usepackage{subcaption}

\numberwithin{equation}{section}

\newtheorem{theorem}{Theorem}[section]
\newtheorem{corollary}[theorem]{Corollary}
\newtheorem{lemma}[theorem]{Lemma}
\newtheorem{proposition}[theorem]{Proposition}
\newtheorem{claim}[theorem]{Claim}
\newtheorem{example}[theorem]{\sl Example}
\newtheorem{definition}[theorem]{Definition}

\theoremstyle{definition}
\newtheorem{remark}[theorem]{Remark}

\newcommand{\gD}{{\Delta}}

\newcommand{\begp}{\begin{proposition}}
\newcommand{\enp}{\end{proposition}}
\newcommand{\begt}{\begin{theorem}}
\newcommand{\ent}{\end{theorem}}
\newcommand{\begl}{\begin{lemma}}
\newcommand{\enl}{\end{lemma}}
\newcommand{\begc}{\begin{corollary}}
\newcommand{\enc}{\end{corollary}}
\newcommand{\begcl}{\begin{claim}}
\newcommand{\encl}{\end{claim}}
\newcommand{\begr}{\begin{remark}}
\newcommand{\enr}{\end{remark}}
\newcommand{\begal}{\begin{algorithm}}
\newcommand{\enal}{\end{algorithm}}
\newcommand{\begd}{\begin{definition}}
\newcommand{\enf}{\end{definition}}
\newcommand{\begx}{\begin{example}}
\newcommand{\enx}{\end{example}}
\newcommand{\bega}{\begin{array}}
\newcommand{\ena}{\end{array}}

\newcommand{\ignore}[1]{}

\def\rompar(#1){\textup(#1\textup)}    

\begin{document}

\newcommand{\tab}[0]{\hspace{.1in}}




\title
{Seeded Graph Matching Via Joint Optimization of Fidelity and Commensurability}
\author{Heather G. Patsolic$^{1}$,
    Sancar Adali$^{2}$, Joshua T. Vogelstein$^{3}$\\
    Youngser Park$^{1}$, Carey E. Priebe$^{1}$, 
    Gongkai Li$^{4}$, Vince Lyzinski$^{5}$  \\
\small{$^{1}$Department of Applied Mathematics and Statistics, Johns Hopkins University, Baltimore, MD, USA}\\
\small{$^{2}$Raytheon BBN Technologies, Cambridge, MA, USA}\\
\small{$^{3}$Department of Biomedical Engineering, Johns Hopkins University, Baltimore, MD, USA}\\
\small{$^{4}$Constellation, Baltimore, MD, USA}\\
\small{$^{5}$Department of Mathematics, University of Maryland
    at College Park, College Park, MD, USA}}
\date{\today}
\maketitle

\begin{abstract}
We present a novel approximate graph matching algorithm that incorporates seeded data into the graph matching paradigm.  Our Joint Optimization of Fidelity and Commensurability (JOFC) algorithm embeds $m$ graphs into a common Euclidean space where the matching inference task can be performed.  Through real and simulated data examples, we demonstrate the versatility of our algorithm in matching graphs with various characteristics---weightedness, directedness, loopiness, many--to--one and many--to--many matchings, and soft seedings.
\end{abstract}

\maketitle
\section{Introduction}

Given $m$ graphs, the graph matching problem (GMP) seeks to find a set of correspondences (i.e., ``matchings")  between the vertex sets that best preserves similar substructures across the graphs.
The graph matching problem has applications across many diverse disciplines including document processing, mathematical biology, network analysis and pattern recognition, to name a few.  Unfortunately, no graph matching algorithm is known to be efficient.  Indeed, even the easier problem of matching isomorphic simple graphs is of famously unknown complexity (see \cite{gandj}).
Because of its practical applicability, there exist numerous approximate graph matching algorithms in the literature; for an excellent survey of the existing literature, see \cite{30ygm}.

When matching across $m$ graphs, often partial correspondences, or
seedings, between the vertices of some pairs of graphs are known. One
cutting-edge algorithm for seeded graph matching, the Seeded Graph
Matching (SGM) algorithm of \cite{FAP} and \cite{sgm2}, leverages the
information contained in seeded vertices to efficiently match graphs
with thousands of vertices, achieving excellent performance with
relatively few seeds. However, as demonstrated in \cite{FAP}, SGM
achieves its optimal performance in the case of highly structured simple
graphs on identical vertex sets.  Although it can be modified to handle
directed, weighted, and other non-simple graphs,
in the presence of these generalizations the performance of the SGM
algorithm deteriorates. 
Moreover, the algorithm cannot currently handle matchings across graphs
that are not one--to--one. 
Often graphs arising from real data contain many of the aforementioned
characteristics, and more robust procedures are needed to effectively
match these graphs.

Herein we present a new seeded graph matching algorithm derived from the Joint
Optimization of Fidelity and Commensurability (JOFC) algorithm of
\cite{JOFC},
extending the preliminary results of \cite{adali2014joint}.
Our algorithm is flexible enough to handle many of the
difficulties
inherent to real data, while simultaneously not sacrificing too much
performance (compared to SGM) when matching across simple graphs.  The
paper is laid out as follows: In Section \ref{S:SGM}, we define the
classical GMP and present the details of the SGM algorithm. 
In Section
\ref{S:pf}, we reformulate the GMP to incorporate non-simple graphs with
potentially different numbers of vertices, and in Sections \ref{S:embed}
-- \ref{S:nonseeds}, we present our JOFC seeded graph matching problem
in detail.
 In Section \ref{S:ex}, we present two simulated and two real
data examples. In the simulated data examples presented in Section
\ref{S:bf}, we note that although the JOFC algorithm is outperformed by
the SGM algorithm on highly structured simple data (Figure
\ref{fig:fig2}), the JOFC algorithm---unlike the SGM algorithm---can
easily handle the case of many--to--one and many--to--many matchings
(Figure \ref{fig:fig3}).
 In Section \ref{S:cel}, we match C. elegans
chemical and electrical connectomes using both the JOFC and SGM
procedures, and in Figure \ref{fig:fig4} we show that our JOFC procedure
significantly outperforms the SGM algorithm in matching across the
connectomes.  We also demonstrate the ability of our JOFC to incorporate
soft-seeded vertices for vertex classification (Figure
\ref{fig:fig5}).
 In Section \ref{S:fb}, we match multiple time series
graphs generated from a zebrafish brain, again demonstrating the
validity of our JOFC algorithm (see Figure \ref{fig:fig6}).

\noindent {\it Note:}  We will define $P(n)$ to be the set of $n$
$\times$ $n$ permutation matrices, and $D(n)$ to be the set of $n$
$\times$ $n$ doubly stochastic matrices. Furthermore, given a graph $G =
(V,E)$ with vertex set $V$ and edge set $E$, by $i\sim_G j$ we mean
there is an edge from vertex $i$ to vertex $j$ in $G$; similarly $i
\not\sim_G j$ indicates that there is not an edge from $i$ to $j$ in
$G$.

\section{The classical graph matching problem and the SGM algorithm}\label{S:SGM}

In its classical form, the two-graph matching problem is as follows:  Given two graphs $G_1$ and $G_2$ on the same vertex set $V$ (with $|V|=n$), we seek to find a bijection $\phi:V\mapsto V$ that minimizes the number of edge disagreements induced by $\phi$; specifically, we seek a bijection $\phi:V\mapsto  V$ that minimizes
\begin{align}
\label{eq:edgedisagree}
d(\phi)=\big|\big\{(i,j)\in V\times V: [i\sim_{G_{1}}j,&\,\phi(i)\nsim_{G_{2}}\phi(j)]\text{ or}[i\nsim_{G_{1}}j,\,\phi(i)\sim_{G_{2}}\phi(j)]\big\}\big|.
\end{align}
Equivalently stated, if the adjacency matrices for $G_1$ and $G_2$ are respectively $A$ and $B$, the problem seeks a permutation matrix $P\in P(n)$ minimizing $\| A-PBP^T\|_F$, where $\|\cdot\|_F$ is the usual matrix Froebenius norm.
If we allow $G_1$ and $G_2$ to be directed, loopy, and weighted, then the classical graph matching problem is equivalent to the quadratic assignment problem, and therefore is known to be NP--hard.  Hence no efficient exact graph matching algorithm is known.

We can generalize the above classical graph matching problem between two graphs to matchings among $m$ graphs:  Given $m$ graphs $G_1$, $G_2,$ $...,$ $G_m$ on the same vertex set $V$ (with $|V|=n$), we seek to find a set of bijections $\Phi=\{\phi^{p,q}\}_{1 \leq p < q \leq m}$, where $\phi^{p,q}:V\mapsto V$ is a mapping between graph $G_p$ and $G_q$, such that the set $\Phi$ minimizes
\begin{align}
\label{eq:edgedisagree2}
d(\Phi)=\sum\limits_{\phi^{p,q} \in \Phi} \big|\big\{(i,j)\in V\times V: [i\sim_{G_{p}}j,&\,\phi^{p,q}(i)\nsim_{G_{q}}\phi^{p,q}(j)]\text{ or}[i\nsim_{G_{p}}j,\,\phi^{p,q}(i)\sim_{G_{q}}\phi^{p,q}(j)]\big\}\big|.
\end{align}
Equivalently, if the adjacency matrix for graph $G_k$ is $A_k$, the problem seeks a set of permutation matrices $\{P_{p,q}\} \subseteq P(n)$ minizing $\sum\limits_{ 1\leq p < q \leq m} \| A_p-P_{p,q}A_qP_{p,q}^T\|_{F}$, where $\|\cdot\|_F$ is the usual matrix Froebenius norm.

Often when matching across graphs, we have access to a partial matching of the vertices in the form of seedings.
If we are given, for each graph $G_k$ a subset $S_{k}\subset V$ of size $s_k$ of the vertices called {\it seeds} and {\it seeding} functions
 $\sigma^{p,q}$ which maps a subset of $S_p$ to a subset of $S_q$, the classical seeded graph matching problem (SGMP) then seeks to minimize (\ref{eq:edgedisagree}) over bijections $\phi^{p,q}:V\mapsto V$ satisfying $\phi^{S_p,S_q}=\sigma^{p,q}$ where $\phi^{S_p,S_q}$ is the restriction of $\phi^{p,q}$ to an appropriate subset of $S_p$.

For the two-graph matching case, the state-of-the-art approximate seeded
graph matching algorithm, the SGM algorithm of \cite{FAP} and
\cite{sgm2}, begins by relaxing the SGMP to minimize $\|A-PBP^T\|_F$
over doubly stochastic matrices $P$ of the form $P=I_s\oplus P'$ with
$P'\in D(n-s)$, where $\oplus$ indicates the direct sum between
matrices.
The algorithm then utilizes Frank-Wolfe methodology to efficiently solve the relaxed problem and finally projects this relaxed solution onto $P(n)$.  The deterioration of the SGM algorithm's performance on non-simple graphs motivates the need for more robust seeded graph matching procedures, such as the JOFC algorithm presented herein.

\begr
It bears noting that there are a multitude of variations on the classical problem in the literature, where different graph attributes give rise to objectives other than minimizing (\ref{eq:edgedisagree}).
For an excellent survey of the existing literature, including many of the current variations on the classical problem, see \cite{30ygm}.  We choose here to focus on the classical problem (\ref{eq:edgedisagree}), as it is closely related to the SGM seeded graph matching algorithm.
\enr

\section{Seeded graph matching via JOFC}
\label{S:JOFC}

We presently approach the seeded graph matching problem via a modification of the Joint Optimization of Fidelity and Commensurability (JOFC) algorithm of \cite{JOFC}, which was originally designed for manifold matching.
Briefly, our algorithm embeds a list of $m$ graphs into a common Euclidean space where our matching inference task can be performed.
The embedding seeks to maximize the information contained both within the connectivity structure of each graph and the across graph relationship provided by the seeding, i.e.\@ we seek to maximize the {\it fidelity} and {\it commensurability} of the embedding.
Once embedded, finding the optimal matching between the vertices then amounts to solving a generalized tensor assignment problem.

We will present our algorithm in its most general form and will note when certain assumptions on our graphs necessarily lead to simplifications.

\subsection{Setup}
\label{S:pf}
\begin{figure}
\hspace{10mm}
\centering
\includegraphics[width=.7\textwidth]{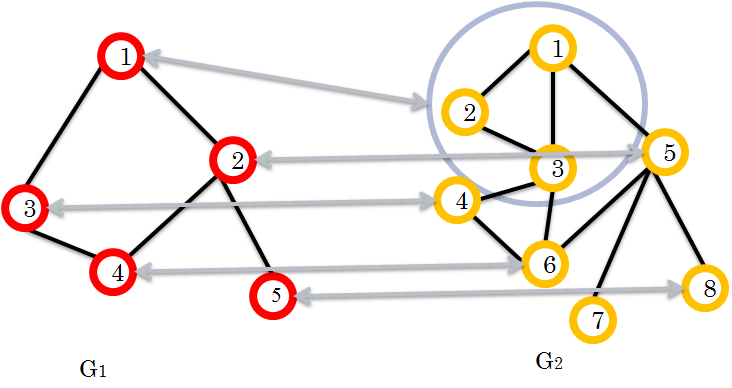}
\caption{For graphs $G_1$ and $G_2$ above, the matching $M_{1,2} \subset V_1\times V_2$ is given via the gray arrows above and is formally defined as
$ M^{1,2} = \{(1,1),(1,2),(1,3),(2,5),(3,4),(4,6),(5,8)\} .$
Note that vertex 1 in $G_1$ is matched to three vertices in $G_2$, namely vertices 1, 2, and 3.  Also vertex $7$ in $G_2$ is not matched to any vertex in $G_1$.}
\label{fig:fig1}
\end{figure}

Let $\{G_k\}_{k=1}^{m}$ be graphs on respective vertex sets $\{V_k:=V(G_k)\}_{k=1}^{m}$ (with each $|V_k|=n_k$).  Without loss of generality, we assume the vertices are labeled $V_k=\{1,2,\ldots,n_k\}$.  As we no longer assume all graphs have the same set of vertices, the graph matching problem as stated in (\ref{eq:edgedisagree}) is not necessarily well-posed.
Rather than reformulating the classical GMP in terms general enough to handle all of the difficulties inherent to real data problems, we choose instead to reformulate our approach to graph matching.
We begin with a set of assumed true matchings $\mathcal{M}=\{M^{p,q}\}_{1 \leq p < q \leq m}$ between every pair of vertex sets $V_p$ and $V_q$, though in the present general setting a matching is simply a subset $M^{p,q}\subset V_p\times V_q$.
If $(u,v)\in M^{p,q}$, then $u$ and $v$ are ``matched" vertices in
graphs $G_p$ and $G_q$, though the precise definition of ``matched" here
is context specific.  In one setting, $u$ and $v$ could be the same
actor in two different communication graphs, while in another setting, $u$
and $v$ could represent the same neuron in two different
neuro-connectome graphs.  In each of our real and simulated data
examples, the true matching is explicit from the context of the
problem.
 Note that although it is often the case that $n_p=n_q$ and
the matching $M^{p,q}$ is a bijection between the vertex sets, our more
general definition allows for multiple vertices in $G_p$ to be matched
to a single vertex or no vertex at all in $G_q$, and vice-versa.  See
Figure \ref{fig:fig1} for an illustrative example of a matching in this
general context.
 Our inference task is then to leverage the
information contained in seeded vertices to estimate the true underlying
matching $\mathcal{M}$.

In this newly reformulated graph matching problem, a {\it seeding} refers to a collection $\mathcal{S}=\{S^{p,q}\}_{1 \leq p < q \leq m}$, where given any two graphs $G_p$ and $G_q$, $S^{p,q}\subset M^{p,q}\subset V_p\times V_q$ with the following property: if $(i,j)\in S^{p,q}$, $(l,h)\in S^{p,q}$ and $(i,h)\notin S^{p,q}$ then $(i,h)\notin M^{p,q}$; i.e. seeded vertices can only be matched to other seeded vertices if the matching is explicitly given by the seeding.
This is an intuitive assumption; indeed it is often natural to assume the full matchedness amongst the seeded vertices is known.  There are certainly applications in which this is not true, and our algorithm can easily be modified to incorporate an incomplete seeding as well.
The vertices in the ordered pairs of $S^{p,q}$ are referred to as {\it seeds}, and we define $S_{p,q}$ as the set of all seeds in $V_p$ that match to seeds in $V_q$.
Furthermore, we define all seeded elements in $V_p$ as $S_{p}$, i.e., $S_{p} := \bigcup\limits_{q: q \ne p} S_{p,q} = \bigcup\limits_{q: q \ne p} \{i : (i, j) \in S^{p,q} \}$.
We will label all the unseeded vertices of $G_p$ and $G_q$ via $U_{p,q}$
and $U_{q,p}$ respectively, so that
$U_{p,q} = \{i \in V_p : \text{ there is no } j \in V_q \text{ so that } 
    (i,j) \in S_{p,q}\}$, and 
similarly for $U_{q,p}\subset V_q$, and define $U_p = V_p \setminus S_p$. 
In addition, we will assume that if 
$(i,j) \in \{v \in V_p : (i,v) \in S_{p,q}\} \times U_{q,p}$ or 
$(i,j) \in U_{p,q} \times \{v \in V_q : (v,j) \in S_{p,q}$, then 
$(i,j)\notin M^{p,q}$.

\noindent{\it Note:}  To simplify later notation, for any subset  $W\subset V_p\times V_q$, we write
$$W(i):=\begin{cases}
\{j\in V_q:(i,j)\in W\}&\text{ if }i\in V_p\\
\{j\in V_p:(j,i)\in W\}&\text{ if }i\in V_q.
\end{cases}$$
In the sequel, we shall also write $|S_{p}|=s_{p}$.

\subsection{Embedding the seeded graph data}
\label{S:embed}
Due to the pathological nature of the graphs we aim to match, performing the matching directly on the graph data proved difficult.
To circumvent this, our algorithm uses multidimensional scaling (MDS) to embed the $m$ graphs into a common Euclidean space where our matching task can more readily occur.

Our embedding begins with $m$ dissimilarity representations $\{\Delta_k\in \mathbb{R}^{n_k\times n_k} : k=1 ,..., m \}$ of $\{G_k\}_{k=1}^{m}$.
We will assume a priori that the dissimilarities have been normalized to be on the same scale.  Ideally, we choose the dissimilarity dependent on the nature of the data, as different dissimilarities will emphasize different aspects of the underlying graph topology.
Although we do not theoretically address the issue of optimally choosing the dissimilarity in the present paper, empirical results have shown that correctly choosing the dissimilarities is essential to the performance of our downstream matching task.  Indeed, in one application to matching neural connectomes of the C. elegans worm, we achieve excellent performance using the weighted DICE dissimilarity of \cite{dice}, a local neighborhood based measure suitable for the sparse structure of the worm brain graphs; see Section \ref{S:cel} for detail.
However, in the simulated Erd\"os-R\'enyi (ER) graph examples of Section \ref{S:bf}, the DICE dissimilarity is not appropriate due to the highly structured nature of the neighborhoods in ER graphs.  We empirically demonstrate a marked performance increase by utilizing a more global dissimilarity, namely the shortest path distance.  Alternately, we could have used diffusion distance, expected commute times, etc.
See
\cite{diffdist}, \cite{dis1}, \cite{dis2} for a wealth of possible dissimilarity representations.



In order for the matching inference task to successfully occur in the embedded space, the embedding must preserve the  information contained both within the connectivity structure of each graph and the between graph relationship given by the matching $\mathcal{M}$.
In essence, the goal of the embedding is simple:  If $(i,j)\in M^{p,q}$, then $i \in G_p$ and $j \in G_q$ are ``matched" vertices and should be embedded close to each other in $\mathbb{R}^d$.  Also, if $u,v\in V_i$ are such that $\gD_i(u,v)$ is small, then $u$ and $v$ are similar vertices in the underlying graph and should also be embedded close to each other in $\mathbb{R}^d$.

Preserving the matching $\mathcal{M}$ in the MDS embedding (or
preserving any available across graph relationship) requires us to
impute an across graph dissimilarity $\delta^{p,q}:V_p\times V_q\mapsto
\mathbb{R}$.  For matched vertices $(i,j)\in M^{p,q}$, it is reasonable
to impute $\delta^{p,q}(i,j)=0$, though for $(i,j)\notin M_{p,q}$, the
imputation is less obvious.
Here we treat these dissimilarities as missing data in the subsequent
MDS procedure.

We do not have access to the full matching $\mathcal{M}$, but the seeding $\mathcal{S}$ provides sufficient information for calculating the imputed $\delta$ amongst the seeded vertices.  For $(i,j)\in S^{p,q}$, we (as before) impute $\delta^{p,q}(i,j)=0$.
For $(i,j)\in {N^{p,q}} :=\{(i,j)\in S_p\times S_q: (i,j)\notin S^{p,q}\}$, we take $\delta^{p,q}(i,j)$ as missing data. Furthermore, rather than incur additional estimation error by imputing the unknown $\delta$ across the unseeded vertices, we also treat these as missing data in our MDS procedure.

We proceed then as follows.  We first embed the seeded vertices and then out-of-sample embed the unseeded vertices using the methodology of \cite{oos}.  With a possible relabeling of the vertices, let the seeded vertices for each graph $G_p$ be denoted as $S_p=\{1,2,\ldots,s_p\}$, so that
$$\Delta_p=\bordermatrix{&S_p&U_p\cr
                S_p&\gD^{(p)}_{1,1} & \gD^{(p)}_{1,2}  \cr
                U_p& \gD^{(p)}_{2,1}  &  \gD^{(p)}_{2,2}}.
                $$
Labeling the embedded vertices of $S_p$ via $\{X^{(p)}_1,X^{(p)}_2,\ldots,X^{(p)}_{s_p}\}$, we define the across-graph squared {\it commensurability error} of the embedding between graph $G_p$ and $G_q$ via
\begin{equation}
\label{eq:comm_pair}
\varepsilon^2_{C, G_p, G_q}:=\sum_{1\leq i\leq s_{p} } \sum_{j\in S^{p,q}(i)} \left(d(X^{(p)}_i,X^{(q)}_j)-\delta^{p,q}(i,j)\right)^2,
\end{equation}
and the total squared across-graphs {\it commensurability error} is given by:
\begin{equation}
\label{eq:comm}
\varepsilon^2_{C}:= \sum\limits_{1 \leq p < q \leq m} \varepsilon^2_{C, G_p, G_q}
\end{equation}
where
$d(\cdot,\cdot)$ is the Euclidean distance between points in $\mathbb{R}^d$.  For $(i,j)\in S^{p,q}$, we impute $\delta^{p,q}(i,j)=0$ and commensurability error between two graphs reduces simply to the squared Euclidean distance between embedded matched vertices.  The commensurability error captures how well the embedding preserves the partial graph matching provided by the seeding.

Note that even if the commensurability of the embedding is small, the embedded points may poorly preserve the original within-graph dissimilarities, which is captured by the fidelity of our embedding.
The within-graph squared {\it fidelity error} of the embedding of $\gD^{(p)}_{1,1}$ is given by
\begin{equation}
\label{eq:fid-p}
\varepsilon^2_{F_p}:=\sum_{1\leq i< j\leq s_p}\left(d(X^{(p)}_i,X^{(p)}_j)-\gD^{(p)}_{1,1}(i,j)\right)^2,
\end{equation}
and the total squared {\it fidelity error} is
\begin{equation}
\label{eq:fid}
\varepsilon^2_{F}:=\sum\limits_{k=1}^{m}\varepsilon^2_{F_k}.
\end{equation}

Closely connected to the fidelity error is the across-graph squared {\it separability error} between two graphs defined via
\begin{equation}
\label{eq:sep_two}
\varepsilon^2_{S, G_p, G_q}:=\sum\limits_{(i,j)\in ( S_p \times S_q \setminus S^{p,q})}\left(d(X^{(p)}_{i},X^{(q)}_{j})-\delta^{p,q}(i,j)\right)^2.
\end{equation}
and the total across-graphs {\it separability error} is
\begin{equation}
\label{eq:sep}
\varepsilon^2_{S}:=\sum\limits_{1 \leq p < q \leq m} \varepsilon^2_{S, G_p, G_q}
\end{equation}
However, since we took $\delta^{p,q}(i, j)$ as missing data in this scenario, we can ingore the separability error for later inference. 

If the errors $\varepsilon^2_{F}$, and $\varepsilon^2_{C}$ are all small and we have embedded the graphs into an appropriate dimension, then we can successfully perform our matching inference task in the target embedding space.
Assuming at present that we know a suitable embedding dimension $d$, we simultaneously control the above errors by jointly embedding the $\sum\limits_{k=1}^{m} s_k$ seeded vertices of the $m$ graphs via the {\it omnibus dissimilarity matrix}
\begin{equation}
\label{eq:omni}
D:=\bordermatrix{&S_1 &S_2 &\ldots &S_m \cr
                 S_1&\gD^{(1)}_{1,1} & \delta^{1,2}      &\ldots     &\delta^{1,m}  \cr
                 S_2&\delta^{2,1}    & \gD^{(2)}_{1,1}   &\ldots     &\delta^{2,m}  \cr
                 \vdots&\vdots       & \vdots            &\vdots     &\vdots        \cr
                 S_m   &\delta^{m,1}  & \delta^{m,2}      &\ldots     &\gD^{(m)}_{1,1}
                 }
\end{equation}

We embed $D$ using the JOFC algorithm of \cite{smacof} for weighted raw stress MDS, where the associated weight matrix is given by
$$W:=\begin{bmatrix}
      J_{s_1}-I_{s_1} & w\cdot O^{1,2} &\ldots   &w\cdot O^{1,m}\\
      w \cdot O^{2,1} & J_{s_2}-I_{s_2} &\ldots   &w \cdot O^{2,m}\\
      \vdots  & \vdots          &\vdots   &\vdots  \\
      w\cdot O^{m,1} &w \cdot O^{m,2}          &\ldots   &J_{s_m}-I_{s_m}\\
      \end{bmatrix} \in \mathbb{R}^{(\sum\limits_{k=1}^{m}s_k) \times (\sum\limits_{k=1}^{m}s_k) },$$
where $J_n$ is an $n \times n$ matrix of all entries being $1$, $w$ is a fixed real number between $0$ and $1$ representing the weight we choose,  and $O^{p,q}$ is a matrix with the same dimensionality as that of $\delta^{p,q}$, whose entries take value $1$ whenever the corresponding entries in $\delta^{p,q}$ is $0$, and take value $0$ whenever the corresponding value is missing in $\delta^{p,q}$ \cite{fastjofc}.

Suppose {\bf X} is some configurations of the $\sum\limits_{k=1}^{m}s_k$
points in $\mathbb{R}^d$. The JOFC algorithm is an iterative procedure
minimizing the cost function
\begin{align}
\label{eq:stress}
\sigma({\bf X}) = \sum\limits_{i<j} W_{i,j}(D(i,j) - d_{i,j}({\bf X}))
\end{align}
over all possible configurations of $\sum\limits_{k=1}^{m} s_k$ points ${\bf X}$ in $\mathbb{R}^d$.  The $W_{i,j}$ are the weights representing our confidence in the dissimilarity
$D(\cdot,\cdot)$ between pairs of vertices.  In our applications, $W$ is designed so that (\ref{eq:stress}) simplifies to
\begin{equation}
\label{eq:stress2}
\sigma({\bf X})=w \varepsilon^2_{F}+(1-w)\varepsilon^2_{C},
\end{equation}
a mixture of the fidelity/separability errors (which capture how well our embedding preserves the original within-graph dissimilarities) and the commensurability error (which captures how well the embedding preserves the partial matching given by the seeds).  This ability to weight the dissimilarities is an essential feature of the JOFC algorithm and is one of the main reasons we have chosen it over more classical multidimensional scaling procedures.  In all of our applications, we have chosen $w=0.8$, and have left the optimal choice of $w$ for future work.

\subsection{Embedding the unseeded vertices}
\label{S:ns}

We next use the procedures outlined in \cite{oos} to out-of-sample embed all the unseeded vertices $U_k$ $k=1, \ldots, m$, obtaining the configuration ${\bf Y}\subset\mathbb{R}^d$ of the $u_k= |U_k|$ unseeded vertices of each graph $G_k$, labeled $\{Y^{(k)}_1,\ldots,Y^{(k)}_{u_k}\}$.
For the out-of-sample embedding, we treat the unknown across-graph dissimilarities involving unseeded vertices as missing data.

The goal of our out-of-sample procedure is simply to preserve the within graph dissimilarities $\Delta^{(k)}_{1,2}$'s.
Indeed, suppose that $(i,j)\in U_p\times U_q$ is such that $(i,j)\in M^{p,q}$.  Ideally, the seeding $S^{p,q}$ will be such that there exists $(u,v)\in S^{p,q}$ such that $\Delta_p(i,u)$ and $\Delta_q(j,v)$ will both be small.
If our two step embedding procedure preserves the seeding $S^{p,q}$ and $\Delta^{(p)}_{1,1}$, $\Delta^{(q)}_{1,1}$, $\Delta^{(p)}_{1,2}$, and $\Delta^{(q)}_{1,2}$ then $d(Y^{(p)}_i, Y^{(q)}_j)$ will be small from a simple triangle inequality argument.
If $(i,j)\in U_p\times U_q$ is such that $(i,j)\notin M^{p,q}$, then the seeding $S^{p,q}$ ideally has the property that there exists $(u,v)\in S^{p,q}$ such that one of $\Delta_p(i,u)$ and $\Delta_q(j,v)$ is small and the other one is large.
If our two step embedding procedure preserves the seeding $S^{p,q}$ and $\delta^{p,q}$, $\Delta^{(p)}_{1,1}$, $\Delta^{(q)}_{1,1}$, $\Delta^{(p)}_{1,2}$, and $\Delta^{(q)}_{1,2}$ then $d(Y^{(p)}_i, Y^{(q)}_j)$ will be large from another simple triangle inequality argument.
Assuming the above, the matching $M$ amongst unseeded vertices will then be preserved under the embedding without the need to impute the unknown $\delta$ across unseeded vertices.


Following \cite{oos}, our embedding procedure then seeks to minimizes the stress function:
\begin{align}
\label{eq:oos}
\sigma({\bf Y})&=\sum_{p=1}^{m}\sum_{i=1}^{s_p} \sum_{j=1}^{u_p}\widetilde w^{p,q}(X^{(p)}_i,Y^{(p)}_j)\left(d(X^{(p)}_i,Y^{(p)}_j)-\gD^{(p)}_{1,2}(i,j)\right)^2\\
\label{eq:oos1}
&+\sum_{p=1}^{m}\sum_{i=1}^{u_p}\sum_{j=1}^{u_p}\widetilde w^{p,q}(Y^{(p)}_i,Y^{(p)}_j)\left(d(Y^{(p)}_i,Y^{(p)}_j)-\gD^{(p)}_{2,2}(i,j)\right)^2
\end{align}
over configurations ${\bf Y}$. Here $\widetilde w^{p,q}(\cdot,\cdot):V_p\times V_q \mapsto\mathbb{R}$ is a weighting function representing our confidence in the computed dissimilarity between pairs of vertices.
In our applications, we have chosen to zero out the weighting function $\widetilde w^{p,q}$ between unseeded vertices within each graph, i.e. we have zeroed out the sums in (\ref{eq:oos1}) from $\sigma({\bf Y})$.  We set the remaining $\widetilde w$'s to be 1.  This is an artifact of our implementation of the out-of-sample embedding procedure, and is not a requirement of our algorithm.
However, in applications where only the 1-neighborhoods of the seeded vertices are known, this would be a naturally enforced constraint.

\subsection{Matching the unseeded vertices}
\label{S:nonseeds}

Supposing the Euclidean distances amongst the unseeded vertices well
preserves the unknown matching $M$ (i.e.\@ if $(i,j)\in U_{p,q} \times
U_{q,p}$ is in $M^{p,q}$, then $d(Y^{(p)}_i,Y^{(q)}_j)$ is small and if
$(i,j)\notin M^{p,q}$, then $d(Y^{(p)}_i,Y^{(q)}_j)$ is large), we
approximate the unknown matching $M$ between unseeded vertices as
follows:
{\it

\begin{enumerate}

\item  Match graph $H=G_1$ and $G_2$ by solving the generalized assignment problem:
\begin{equation}
\min_{\widehat{M^{1,2}}\subset U^{1,2}}\sum_{(i,j)\in \widehat{M^{1,2}}} d(Y^{(1)}_i,Y^{(2)}_j)
\end{equation}
To avoid trivial solutions, we impose the further restriction that $|\widehat{M^{1,2}}(j)|>0$ for all $j\in U_1$.  We do allow vertices in $U_2$ to be unmatched to any vertices in $U_1$

\item  Averaging the matched graphs we get from $G_1$ and $G_2$ to get a
    new graph $H$, that is, if $k_1$ vertices of graph $G_1$ are matched
    with $k_2$ vertices of graph $G_2$, we construct the corresponding
    vertex in $H$ as the average of these $k_1 + k_2$ vertices.

\item  Match graph $G_3$ to this new graph $H$ while enforcing the consistency. That is, if vertex $i$ in $G_1$ matches to both vetex $j$ in $G_2$ and vertex $l$ of $G_3$ in the seeding, then $j$ and $l$ should also match.

\item  Get the new $H$ by taking the average of $H$ and $G_3$ after maching, as in ii.

\item  Repeat this process for every graph $G_k$, $k=4, \ldots, m.$
\end{enumerate}
}

Thus, we can match every $G_k$ to the rest of the graphs while keep the
seeding consistency. The generalized assignment problem is known to be
NP-hard, see \cite{gap2} for background.  However, there are many good
polynomial-time approximation algorithms in the literature, see for example
\cite{gap1}, which we use in our examples.

\section{Demonstrations and Examples}
\label{S:ex}
We will demonstrate the effectiveness of our algorithm by means of a
simple (but illustrative) simulation and two real data experiments which
serve to demonstrate the flexibility inherent to our algorithm.  We
compare the performance of our algorithm with that of the present
state-of-the-art seeded graph matching algorithm (SGM) of
\cite{FAP}, while also pushing the boundary of the state-of-the-art and
applying our algorithm in the settings where SGM breaks down; namely in
the presence of weightedness, directedness, multiple edges,
soft-seeding, and many--to--one and many--to--many matchings.

In the case of matching only two graphs, letting $\widehat{M}$ be our algorithm's approximation of the true matching $M$, we measure our performance via the {\it matched ratio} of the remaining unseeded vertices,
\begin{equation}
\label{matchrat}
\delta^{(s_1+s_2)}=
\frac{\left|\{i\in U_1:\widehat{M}(i) = M(i)\}\right|+\left|\{j\in U_2:\widehat{M}(j) = M(j)\}\right|}{u_1+u_2}.
\end{equation}

We measure the performance of graph matching algorithms by calculating the fraction, $R_m$, of the unseeded vertices correctly matched across the graphs.  In the case where $s_1=s_2=m$ and $n_1=n_2=n$, we calculate
\begin{equation}
\label{matchrat2}
R_m=\frac{|\{i\in U_1:\widehat{M}(i) = M(i)\}|}{n-m}.
\end{equation}
When $s_1\neq s_2$ and $n_1\neq n_2$, $R_{s_1,s_2}$ is calculated via
$$R_{s_1,s_2}=\frac{|\{i\in U_1:\widehat{M}(i) = M(i)\}|+|\{i\in
        U_2:\widehat{M}(i) = M(i)\}|}{u_1+u_2}.$$
Note that the number of unseeded vertices to match decreases as the number of seeded vertices increases.  In all examples, we show how increasing the number of seeded vertices from 0 to some substantive fraction of the total number of vertices significantly increases our relative performance in correctly matching the unseeded vertices.
\begin{figure}
\label{fig2}
\begin{subfigure}{.16\linewidth}
\centering
\includegraphics[width=3\textwidth]{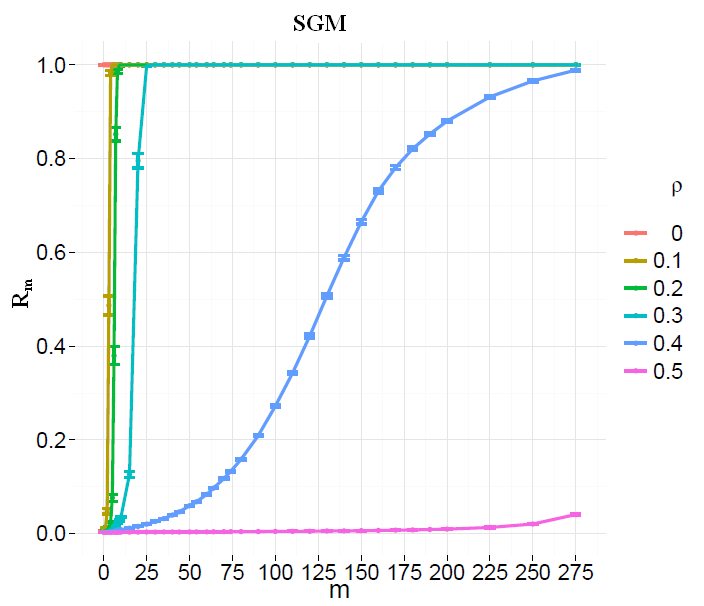}
\caption{}
\label{fig:sub1}
\end{subfigure}
\hspace{60mm}
\begin{subfigure}{.17\linewidth}
\centering
\includegraphics[width=2.8\textwidth, ]{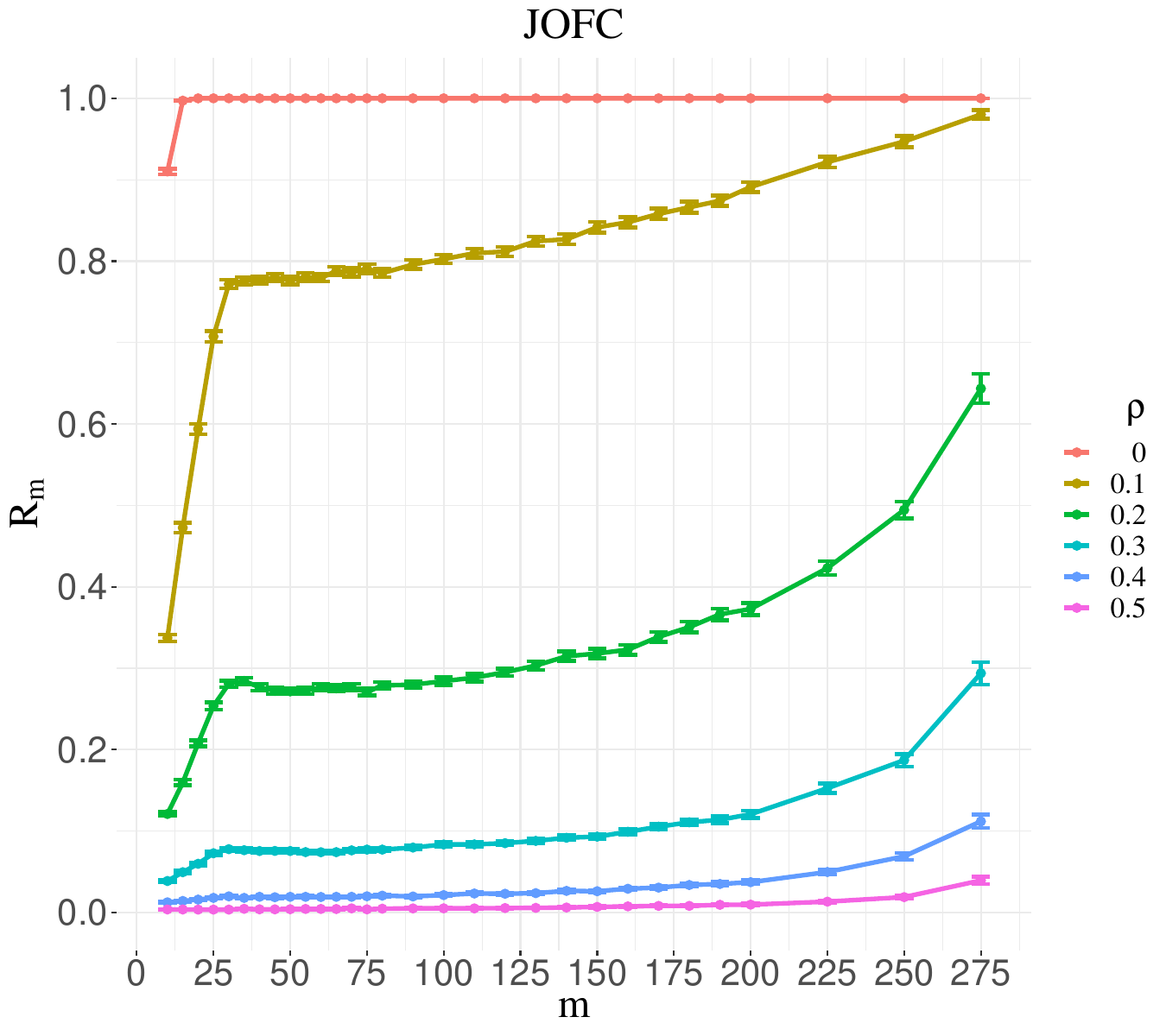}
\caption{}
\label{fig:sub3}
\end{subfigure}
\caption{We plot the matched ratio $R_m\pm 2\, s.e.$ when matching
    across two bit-flipped ER$(300,1/2)$ random graphs for seeds ranging
    from $m=0$ to $m=275$ and bit flip parameter
    $\rho=0,0.1,\ldots,0.5$.  In (a), we show the performance of the SGM
    algorithm of \cite{FAP}; in (b) 
    we plot the performance using our JOFC algorithm
    using shortest path dissimilarities. 
    In each example, we ran 240 MC
    replicates for each combination of $m$ and $\rho$.}
\label{fig:fig2}
\end{figure}

\subsection{The bit-flip model}
\label{S:bf}
We begin with an illustrative simulated data experiment which
simultaneously serves to highlight both the strengths and weaknesses of
our algorithm relative to the cutting edge SGM algorithm.  Let $G_1\sim
ER(n,p)$, a graph on $n$ vertices where each pair of vertices
independently form an edge with probability $p$.  We create a new graph,
$G_2$, by ``flipping bits" in $G_1$ according to the perturbation
parameter $\rho$ as follows: if $u\sim_{G_1}\!v$ then $u\sim_{G_2}\!v$
with probability $(1-\rho)$ and $u\nsim_{G_2}\!v$ with probability
$\rho$; if $u\nsim_{G_1}\!v$ then $u\nsim_{G_2}\!v$ with probability
$(1-\rho)$ and $u\sim_{G_2}\!v$ with probability $\rho$.  Note that if
$\rho=0$ the graphs are identical, and if $\rho=0.5$ the graphs are
independent.%

We consider $n=300$, $p=1/2$ and show the performance of SGM, as well as
the performance of our JOFC algorithm, for varying $m$ and
$\rho$.  We increase $m$ from $0$ to $275$ by increments of 25 and
$\rho$ from $0$ to $0.5$ by increments of $0.1$.  Our JOFC algorithm is
run with the shortest path dissimilarity of \cite{dis2}. 
Note that, using a different dissimilarity measure for the
embedding, such as the weighted DICE dissimilarity of \cite{dice}, can
lead to degredation of performance.
While important, we do not investigate a data-driven heuristic for choosing the
dissimilarity in the present paper. 
We plan to study this further in future work.

In the JOFC implementation, 
for all bit-flip parameters $\rho$, we see a general pattern of increase in
performance when the number of seeds $m$ is increased.  As expected in
this highly
structured simulated data example, SGM performs better than JOFC.
Indeed, SGM achieves its optimal performance in the present
Erd\"os-R\'enyi setting, as shown in \cite{sgm2}.  In the cases where
the data is highly structured and clean, we do not recommend our JOFC
procedure.  It is more appropriate for weighted, directed, loopy, lossy
graphs; i.e.\@ it is more appropriate for real data.  In the real data
examples that follow, we see our algorithm outperform the SGM algorithm.

\begin{figure}[h]
\hspace{20mm}
\centering
\includegraphics[width=.6\textwidth]{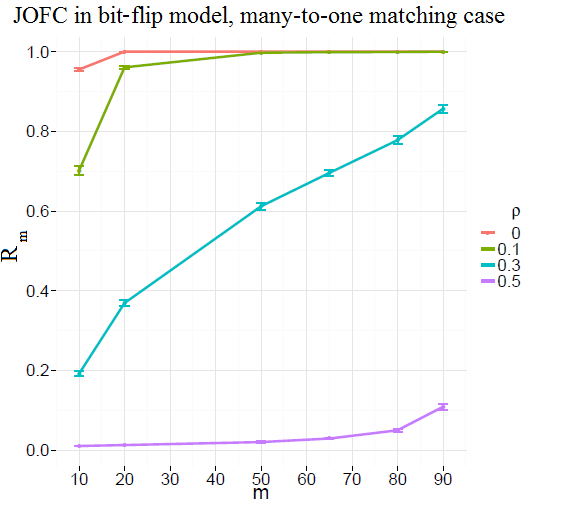}
\caption{Many--to--one matching for $G_1\sim$ER$(100,0,5)$ and $G_2$.
    We create $G_2$ by cloning each vertex in $G_1$ a Geo(.2) number of
    times (with at most 10 cloned vertices created for each vertex) and
    then bit-flipping the cloned graph.  We plot the matched ratio
    $R_m\pm 2\,s.e.$ for 240 MC replicates, $m=10,\,20,\,50,\,65,\,80,\,90$ 
    and $\rho=0,\,0.1,\,0.3,\,0.5$. }
\label{fig:fig3}
\end{figure}

In this simple bit-flip model, we are still able to demonstrate the
flexibility of our algorithm.  While a single iteration of SGM is not
designed for many-to-one matchings,
our JOFC algorithm can handle this difficulty in
stride.  To demonstrate this, we consider the case of many--to--one
matchings in the present bit-flip model.  We begin with
$G_1\sim$ER(100,0.5), and for each vertex create a Geometric(.2) number
of identical vertices in $G_1$ (with at most 10 copies made per vertex).
In the process, we create a new graph $\widetilde G_1$. Here $G_2$ is
the bit-flipped version of $\widetilde G_1$. We then match $G_1$ to
$G_2$; i.e.\@ we seek to match each vertex $i\in V_1$ to its copies in
$G_2$. We measure performance by looking at the ratio of vertices in
$G_2$ matched correctly with the corresponding vertex in $G_1$ for
varying levels of $m,$ the number of seeded vertices.  When we seed here,
for each of the $m$ seeds in $G_1$ we include all matched vertices in
$G_2$ as seeds as well.  The results are summarized in Figure
\ref{fig:fig3}.  Again, note the increased performance as more seeds are
incorporated for all values of the bit--flip parameter and the decreased
performance as the bit--flip parameter is increased.

\subsection{Matching C.\! elegans connectomes}
\label{S:cel}
The Caenorhabditis elegans (abbreviated C.\! elegans) roundworm has been
extensively studied, and is particularly useful due to its simple
nervous system.  The nervous system is believed to be composed of the
same 302 labeled neurons for each organism, with 279 neurons making
synapses with other neurons.  These neural connectomes are mapped in
\cite{vcel}.
There are two types of connections between neurons: chemical (chemical
synapses) and electrical (junction potentials).  We wish to match the
chemical connectome graph $G_c$ and the electrical connectome graph
$G_e$ in order to investigate the extent to which the connectivity
structure alone can be used to identify individual neurons across the
two connectomes.  Here we are considering hermaphroditic worms.  Hence
both $G_c$ and $G_e$ are weighted; $G_e$ is undirected; $G_c$ is
directed; $G_e$ has self-edges, and $G_c$ does not.  Both graphs are
sparse: $G_e$ has 514 undirected edges out of $\binom{279}{2}$ possible
unordered neuron pairs; $G_c$ has 2194 directed edges out of 279$\cdot$
278 possible ordered  neuron pairs.  Before matching the two graphs, we
remove the isolates from each of the individual connectomes, leaving 253
vertices to be matched in each graph.

\begin{figure*}[h]
\centering
\includegraphics[width=.75\textwidth]{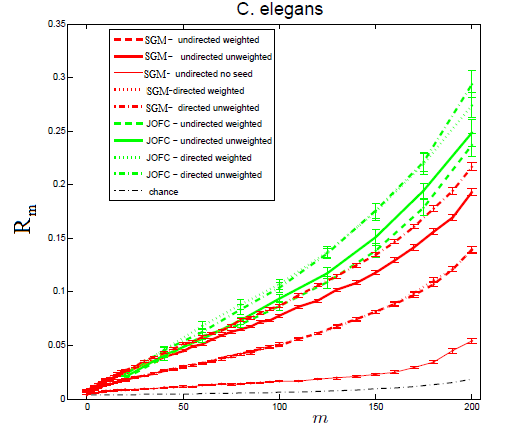}
\caption{Plotting the matched ratio $R_m\pm 2\,s.e.$ for matching the
    $253$ vertex chemical and electrical C. elegans connectomes for seed
    values $m$ ranging from $0$ to $200$.
    We show the performance of the JOFC and SGM algorithms for matching
    the graphs for all combinations of with/without directness and
    with/without edge weights.  JOFC is plotted in green, SGM in red,
    and chance in black.  Note that JOFC for each combination of
    with/without directedness and with/without edge weights
    significantly outperforms the best SGM combination (directed and
    unweighted).  For each combination of $m$ and $\rho$, we ran 100 MC
    replicates.
}
\label{fig:fig4}
\end{figure*}
In Figure \ref{fig:fig4}, we compare the performance of our JOFC
algorithm (utilizing the weighted DICE dissimilarity of \cite{dice})
with the performance of the SGM algorithm in matching across the two
graphs.
For each of SGM and JOFC,
we consider matching with/without edge directions and with/without edge weights.  We see that
best performance is obtained with JOFC, either in the directed unweighted graph case or the directed weighted graph case.  As
expected, performance improves when incorporating more seeds.  For instance, with $m = 50$ seeds, JOFC run on the directed weighted graphs has $R_m\approx0.05$ (chance is $1/203<0.005$) while with $100$
seeds JOFC run on the directed weighted graphs matches the remaining $253-m = 153$ vertices with
$R_m\approx 0.10$ (chance is $1/153 < 0.01$).  Note
that for $m = 100$, JOFC run on the directed weighted graphs matching either $G_c$ to $G_c$ or $G_e$
to $G_e$ is nearly perfect ($R_m > 0.97$ for both cases).

This demonstrates conclusively that there is statistically significant signal in the connectivity structure alone for matching individual neurons across the two connectomes.
The implications for understanding the relationship between neuron connectivity and the information processing properties of the connectome are profound: (i) had the
matching been essentially perfect, the conclusion would have been that one could consider just one
(either one) of the two graphs with little loss of information; (ii) had the matching been essentially
chance, the conclusion would have been that one must consider both graphs, but that they could
be considered separately; (iii) in fact, our results demonstrate that optimal inference regarding the
information processing properties of the connectome must proceed in the joint space.
The results presented in Figure \ref{fig:fig4} demonstrate that seeded matching of $G_c$
to $G_e$ does indeed extract statistically significant signal for identifying individual neurons
across the two connectomes from the connectivity structure alone.

We next demonstrate the potential for our JOFC algorithm to be used for vertex classification.  We consider an experiment
on a collection of neurons categorized as IL and OL (labial neurons), and RI and RM (ring neurons); with the number of non-isolate vertices in each class being $n_{IL} = 8$, $n_{OL} = 6$, $n_{RI} = 18$, and $n_{RM} = 15$.  The total number of neurons under consideration in these four categories is 47.

\begin{figure*}[h]
\begin{subfigure}{.25\linewidth}
\centering
\includegraphics[trim=0cm 0cm 0cm 2cm, clip=true, width=2\textwidth]{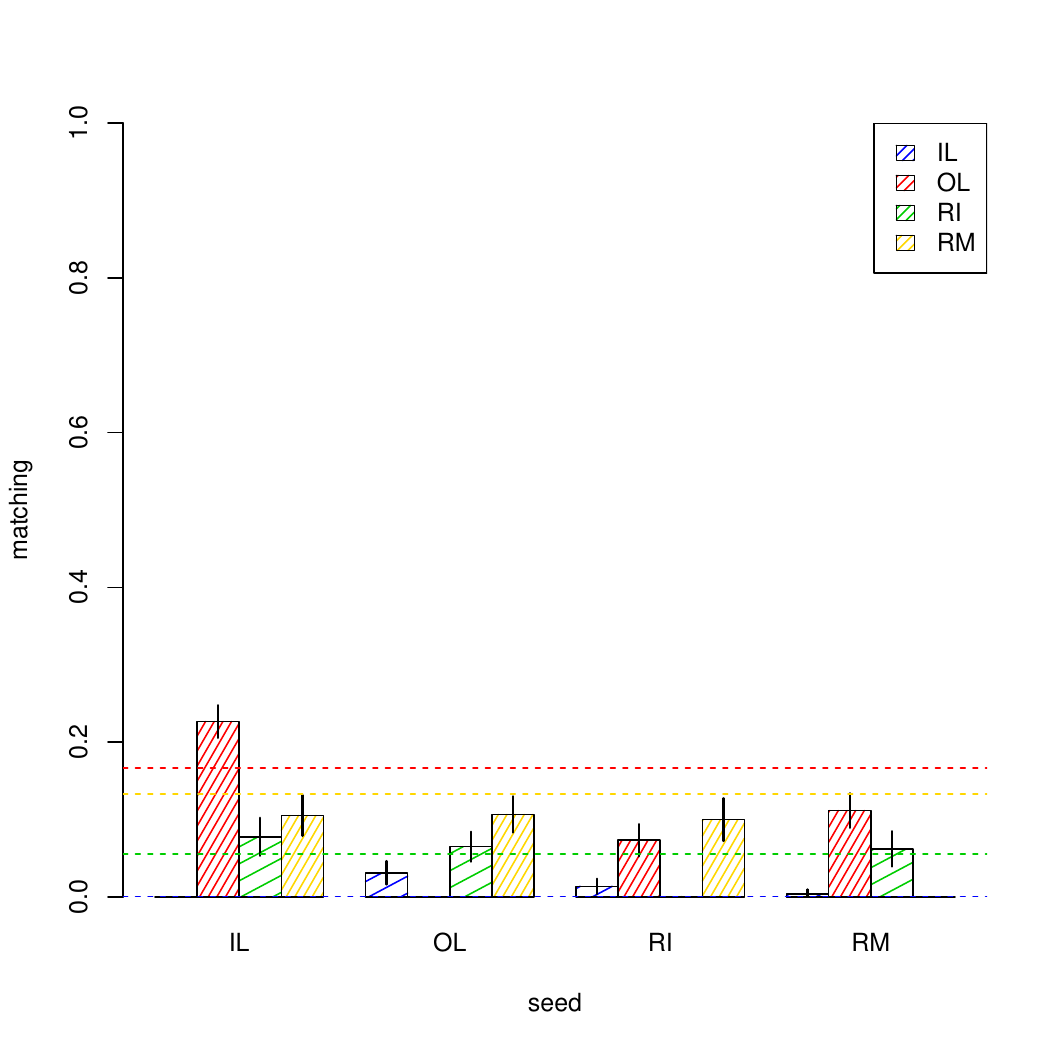}
\caption{}
\label{fig:sub5}
\end{subfigure}%
\hspace{35mm}
\begin{subfigure}{.25\linewidth}
\centering
\includegraphics[trim=0cm 0cm 0cm 2cm, clip=true, width=2\textwidth,]{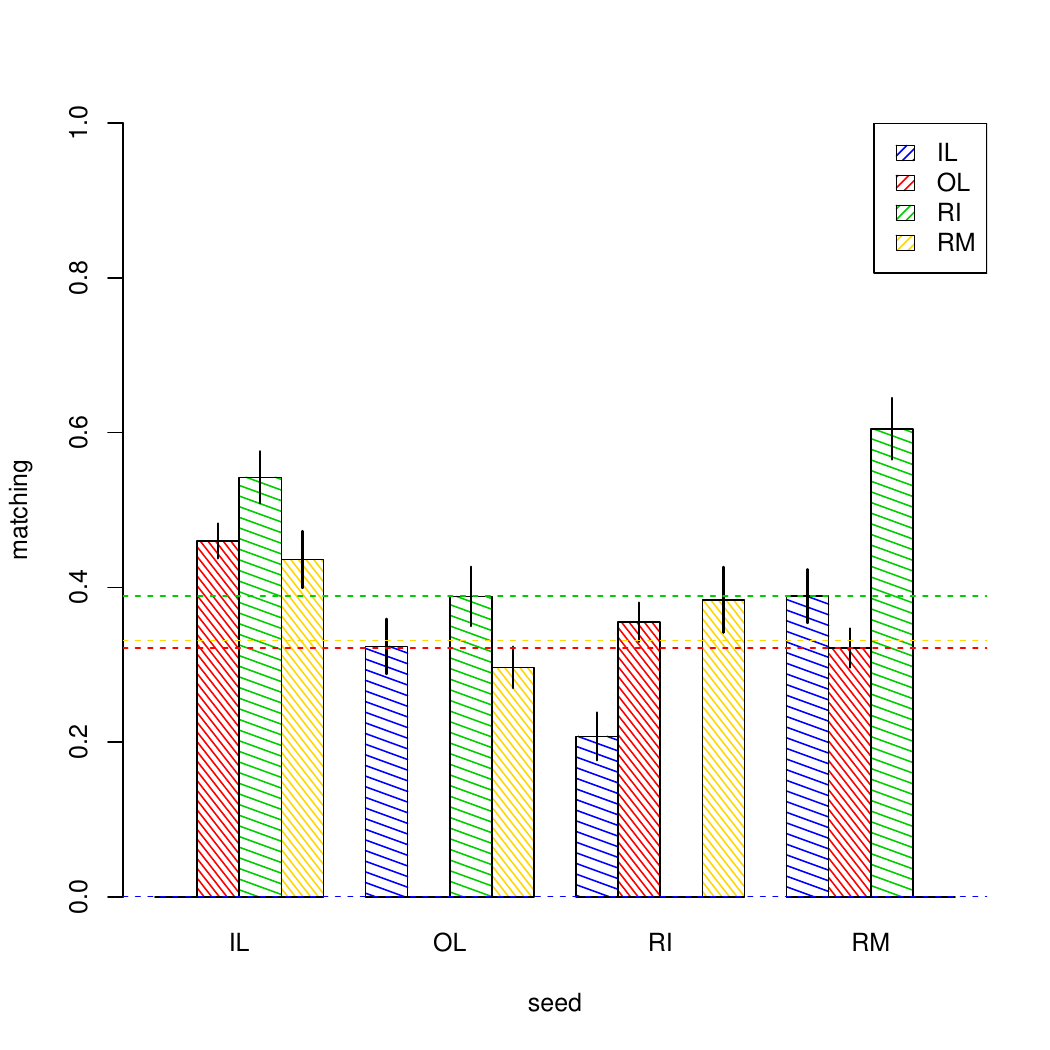}
\caption{}
\label{fig:sub6}
\end{subfigure}
\caption{Results for our demonstration of JOFC's classification
    potential.  In (a) we plot the fraction of categorized vertices
    matched correctly across the graphs; and in (b) we plot the fraction
    of categorized vertices matched to the correct category across the
    graphs.  In both cases, the horizontal dotted lines represent the
    case I results (when the seeded vertices are the 206 uncategorized
    vertices). Results for case II are presented in the bar graphs
    in figures (a) and (b).  For example, the three bars above IL in
    figure (a) (resp. in (b)) represent the fraction of each of the
    other three categories matched correctly (resp. matched to the
    correct category) when the seeds are the $n_{IL}$ IL vertices and
    $206-n_{IL}$ random chosen uncategorized vertices.  Again, we used
    100 MC replicates for each seeding level.}
\label{fig:fig5}
\end{figure*}

We employ $m=253-47=206$ seeds not in these four categories.  We first
(case I) let all 206 vertices not in categories IL, OL, RI and RM be our
seeded vertices, and we seek to correctly classify the 47 remaining
vertices into their proper category.  We measure the number of the 47
vertices matched correctly across the graphs and also measure the number
matched to a vertex of the correct category.  Second (case II), for each
of the four categories $c\in\{ IL;OL;RI;RM\}$ in turn, the $m$ seeds are
chosen to be all the neurons in category $c$ together with $m-n_c$ seeds
chosen randomly from amongst the 206 neurons not in these four
categories.  Again, we measure the number of the $47-n_c$ vertices
matched correctly across the graphs (Figure \ref{fig:fig5}a) and
measure the number matched to a vertex of the correct category (Figure
\ref{fig:fig5}b).  Note the effect that the different choices of
seedings has on the matching performance.  Indeed, ``informative" seeds
can greatly increase the matching performance in our algorithm, and in
future work we plan to investigate heuristics for optimizing the
information in our selected seeds.  The results are summarized in Figure
\ref{fig:fig5}.

\subsection{Matching zebrafish brains}
\label{S:fb}

In \cite{fishbrain}, the authors present their research involving a time
series dataset for a zebrafish 
brain using Light-Field Deconvolution Microscopy and pan-neuronal
expression of GCaMP, which is a fluorescent calcium indicator that
serves as a proxy for neuronal activity \cite{fishbrain, fastjofc}. 
From this image data, we generate a time series of 20 graphs $\{ G^{(t)}
\}$ indexed by time $t = 1, 2, \ldots, 20$ in which each graph $G^{(t)}$ is
generated on the same fixed 469 neurons (vertices) and the vertices $i$
and $j$ are adjacent if and only if the neuronal activities between
neuron $i$ and $j$ cross some threhold at time $t$. That is, for each
graph $G^{(t)}$, there is an associated adjacency matrix $A^{(t)}$, such
that $A_{i,j} ^{(t)} = 1$ if there is an edge between vertex $i$ and
$j$.  We have, therefore, a complete true matching between vertices of
all 20 graphs. Furthermore, the $i$th row of the adjacency matrix
$A^{(t)}$ completely characterizes the status of $i$th neuron in
relation to other neurons at time $t$ in the brain. We will demontrate
our JOFC scheme by comparing the matchings of the unseeded vertices with
this known ``ground truth''.
Initial change point detection analysis reveals that there was an
anomaly occuring at $t=13.$ 
For our purpose, we select graphs $G^{(8)}$,
$G^{(9)}$, $G^{(10)}$, $G^{(11)}$, $G^{(12)}$, $G^{(13)}$ and choose for
each graph, the (same) first 300 vertices as seeds and out-of-sample the
remaining 169 vertices for matching. 
As mentioned before, the choice of distance measure plays a crucial role
in the graph matching problem; here we pick $1$ minus the Jaccard index
as our notion of distance. For details about the Jaccard index, see
\cite{jaccard}

\begin{figure*}[h]
\centering
\includegraphics[width=0.32\textwidth]{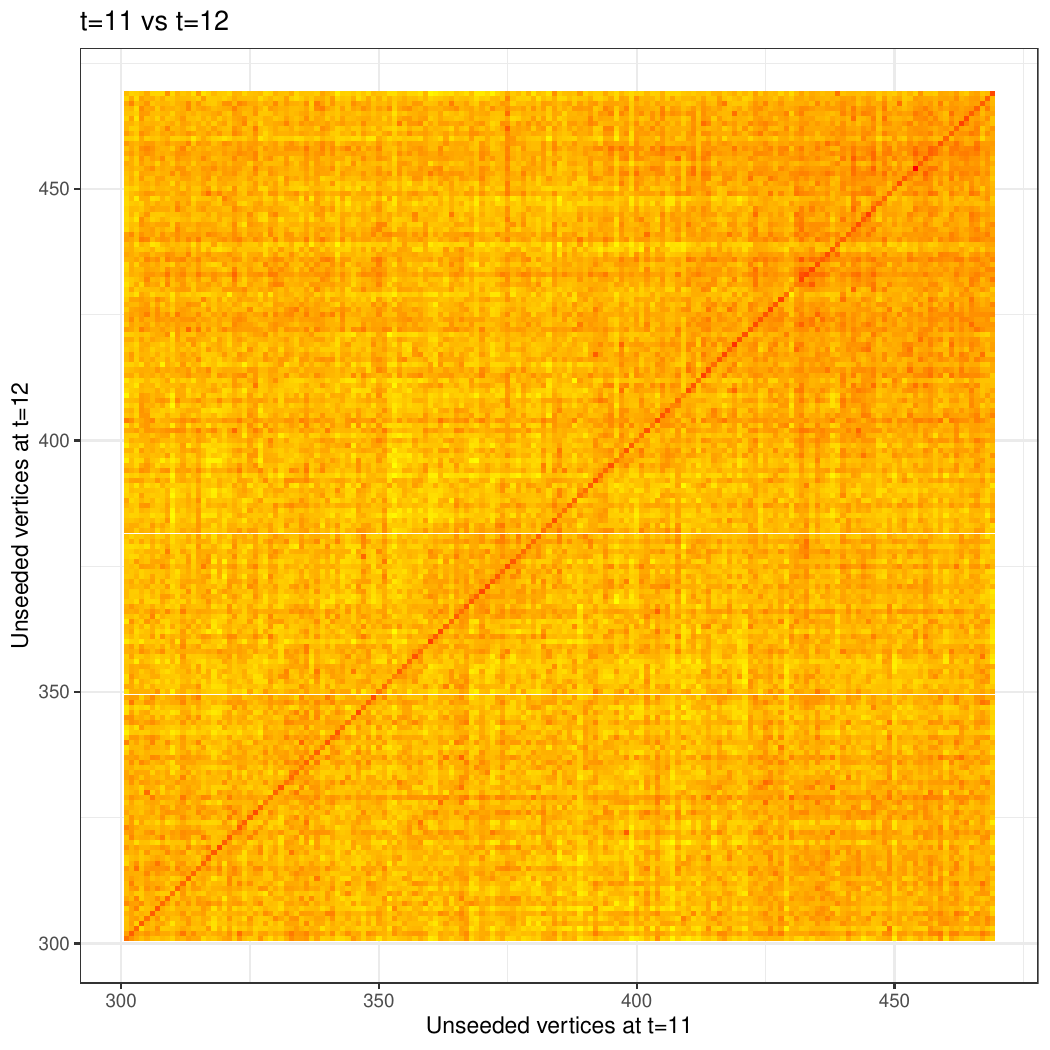}
\includegraphics[width=0.32\textwidth]{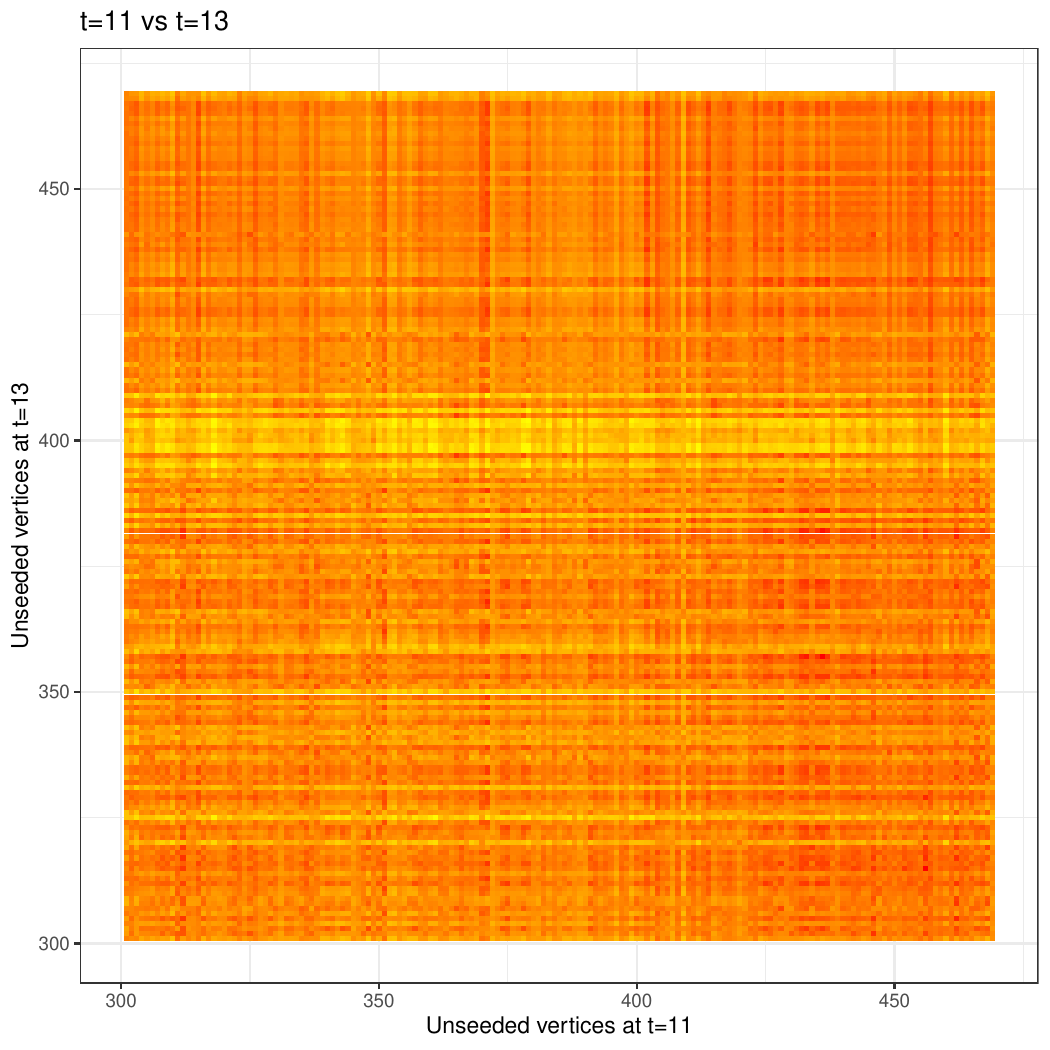}
\includegraphics[width=0.32\textwidth]{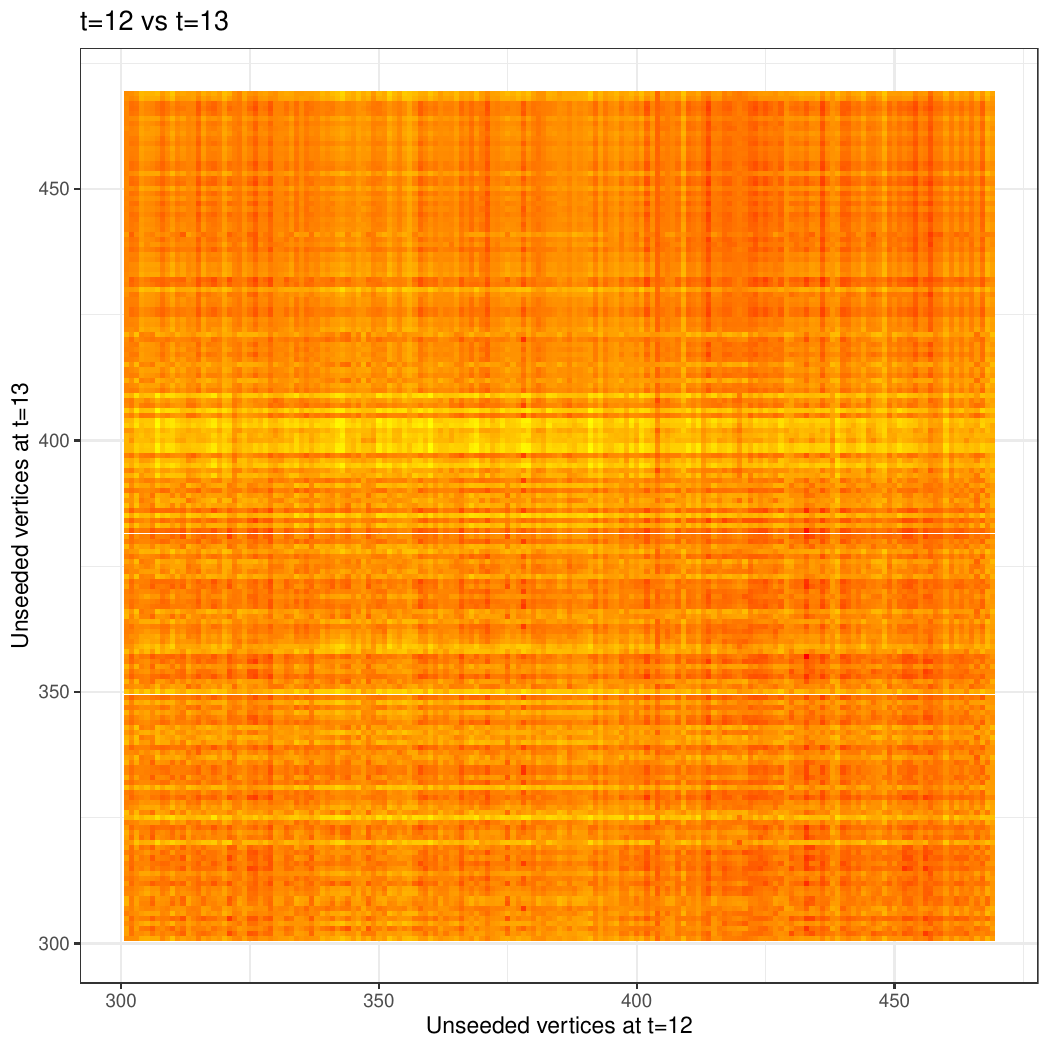}
\caption{The pairwise distance heat map of the unseeded vertices from 3
    graphs, at times $t$=11, $t$=12, and $t$=13. The darker colors
    correspond to smaller distances, the $x$-axis represents the
    unseeded vertices $301$--$469$ in one graph, and the $y$-axis
    represents the unseeded vertices $301$--$469$ in another graph. 
    Since we know the true matchings between all vertices, 
    we should ideally see a dark line
    from lower left to upper right between $t=11$ and $t=12$, which can
    be readily seen. However, since there is an anomaly at $t=13,$ the
    unseeded vertices at $t=13$ will not be matched correctly, which
    also can be seen.}
\label{fig:fig6}
\end{figure*}

We first consider matching graphs $G^{(11)},$ $G^{(12)}$, and $G^{(13)}$
pairwise. 
To match the unseeded vertices, we calculate the pairwise distances
between all pairs of unseeded vertices from different graphs and rank
them in increasing order after the embedding. 
For example, between graphs $G^{(11)}$ and $G^{(12)}$, the $k$th
unseeded vertex in $G^{(11)}$ will be more likely to be matched with the
unseeded vertex in graph $G^{(12)}$ that is the closest to it.
In this manner, not only do we have a matching, but we also get a list of likely matchings between unseeded vertices.
As the anomaly occurs at time $t=13$, we expect that the
output of JOFC should favor the true matching for $G^{(11)}$ and
$G^{(12)}$, as is indicated by the dark diagonal line in the left-most
panel of Figure \ref{fig:fig6}. On the other hand, matching $G^{(11)}$
to $G^{(13)}$ and $G^{(12)}$ to $G^{(13)}$ should not recover the true
alignment, as indicated in the middle and right-most panels of Figure
\ref{fig:fig6}. This example demonstrates that when the true alignment
is known, the JOFC algorithm we propose can be used to detect at which
time-point an anomaly occurs as the matchability decreases.



\begin{figure*}[h]
\begin{subfigure}{.48\linewidth}
\centering
\includegraphics[trim=1cm 2cm 1cm 2cm,clip=true,width=1\textwidth]{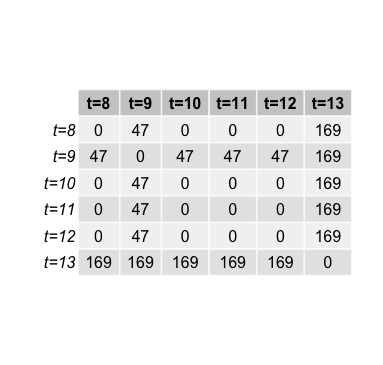}
\caption{We use graph $G^{(8)}$ as the first graph to merge.}
\label{fig:sub8}
\end{subfigure}%
\hspace{5mm}
\begin{subfigure}{.48\linewidth}
\centering
\includegraphics[trim=1cm 2cm 1cm 2cm, clip=true, width=1\textwidth,]{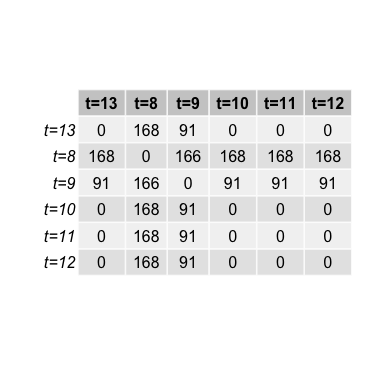}
\caption{We use graph $G^{(13)}$ as the first graph to merge.}
\label{fig:sub9}
\end{subfigure}
\caption{Confusion matrix for graphs at $t=8,9,10,11,12,13$.}
\label{fig:fig7}
\end{figure*}

Next we apply JOFC to the task of aligning all $6$ afore-mentioned graphs for
time-points $t=8,9,\ldots,13$. Since the anomaly occurs at $t=13$, the
first $5$ graphs should match almost perfectly with each other while
$G^{(13)}$ will not be correctly matched to any one of the first $5$
graphs. However, since the algorithm begins by matching two of the
graphs, and subsequently matching a subsequent graph to the average of
the previously aligned graphs, the order in which the matchings occur
matters. In Figure \ref{fig:fig7}, we present a confusion matrix showing
the number of vertices incorrectly labeled at subsequent iterations when
the graphs are matched in orders $t=(8,9,10,11,12,13)$ (left) and 
$t=(13,8,9,10,11,12)$ (right). As might be expected, the confusion
matrices show that the anomaly at time $t=13$ results in less accuracy
of the matching when the matching begins with the anomaly.

\section{Discussion}
The types of graphs common to real data applications are often very far
from well structured random graph models like the Erd\"os-R\'enyi graph
model and stochastic blockmodel.
 
To be readily applicable, graph matching
algorithms need to be robust to the presence of weightedness,
directedness, loopiness, many--to--one and many--to--none matchings,
etc; i.e.\@ they need to be robust to the difficulties inherent to real
data.  Our JOFC approach to graph matching, embedding the graphs into a
common Euclidean space and matching across embedded graphs, is flexible
enough to handle many of the pathologies inherent to real data while
simultaneously not sacrificing too much performance when matching
simulated idealized graphs.  We demonstrate the effectiveness of our
algorithm on a variety of real data examples, for which our JOFC
approach performs significantly better than the cutting edge SGM
procedure.

In presenting our algorithm, we noted many directions for future
research.  Figure \ref{fig:fig2} points to the potential for dramatic
performance increase when choosing an appropriate dissimilarity
for the graph data.  In future work, we plan on pursuing this question
further, seeking principles for dissimilarity choice based upon the
underlying graph topology.   In Figure \ref{fig:fig5}, we see the effect
of well chosen seeds on our matching performance.  In \cite{lsgm}, the
authors present a heuristic for active seed selection in the SGM
procedure, and we are working towards a similar result for our JOFC
algorithm.  Additionally, the reliance of our algorithm on
missing-data MDS approaches greatly limits its scalability to big data
graphs.  We are working towards a scalable missing-data MDS procedure
that is essential for large scale application of our JOFC procedure.
Lastly, we are working towards a theoretically justified dimension
selection procedure which combines our automated approach with the
spectral approaches of \cite{scree}.  In our applications, dramatic
performance is possible when embedding to an appropriate dimension.

\section{Acknowledgments}
This work is partially supported by a
National Security Science and Engineering Faculty Fellowship (NSSEFF)
and the
Johns Hopkins University Human Language Technology Center of Excellence
(JHU HLT COE).
This material is based on research sponsored by the XDATA program of the
Defense Advanced Research Projects Agency (DARPA) administered through
Air Force Research Laboratory contract FA8750-12-2-0303 and the Air
Force Research Laboratory and DARPA, under agreement number FA8750-18-2-0035. 
The U.S. Government is authorized to reproduce and distribute reprints
for Governmental purposes notwithstanding any copyright notation
thereon.The views and conclusions contained herein are those of the
authors and should not be interpreted as necessarily representing the
official policies or endorsements, either expressed or implied, of the
Air Force Research Laboratory and DARPA, or the U.S. Government.
\newpage

\bibliographystyle{plain}
\bibliography{refbib}

\end{document}